\documentclass[10pt,twocolumn,letterpaper]{article}

\usepackage{cvpr}              %

\usepackage{graphicx}
\usepackage{amsmath}
\usepackage{amssymb}
\usepackage{booktabs}
\usepackage{multirow}
\usepackage{xcolor}

\usepackage[pagebackref,breaklinks,colorlinks]{hyperref}
\usepackage{xcolor}
\usepackage{colortbl}

\usepackage[capitalize]{cleveref}
\crefname{section}{Sec.}{Secs.}
\Crefname{section}{Section}{Sections}
\Crefname{table}{Table}{Tables}
\crefname{table}{Tab.}{Tabs.}

\newcommand{\Tref}[1]{Table~\ref{#1}}

\newcommand{\Fref}[1]{Fig.~\ref{#1}}
\newcommand{\Sref}[1]{Section~\ref{#1}}

\definecolor{mygreen}{RGB}{20, 190, 10}
\definecolor{myorange}{RGB}{250, 86, 36}
\definecolor{myblue}{RGB}{100, 100, 255}

\definecolor{LightGray}{RGB}{215,215,215}
\definecolor{LightLightGray}{RGB}{234,234,234}

\usepackage{setspace}
\AtBeginDocument{%
  \addtolength\abovedisplayskip{-0.4\baselineskip}%
  \addtolength\belowdisplayskip{-0.4\baselineskip}%
  \addtolength\abovedisplayshortskip{-0.4\baselineskip}%
  \addtolength\belowdisplayshortskip{-0.4\baselineskip}%
}

\def\cvprPaperID{****} %
\def\confName{CVPR}
\def\confYear{2023}

\begin{document}

\title{Improving Commonsense in Vision-Language Models via \\ Knowledge Graph Riddles}

\author{ \hspace{-2.5em} Shuquan Ye\textsuperscript{1} \quad Yujia Xie\textsuperscript{2} \quad Dongdong Chen\textsuperscript{2} \quad Yichong Xu\textsuperscript{2} \\
Lu Yuan\textsuperscript{2} \quad Chenguang Zhu\textsuperscript{2} \quad Jing Liao\textsuperscript{1}\thanks{Jing Liao is the corresponding author.} \\
\\
\hspace{3em} \textsuperscript{1} City University of Hong Kong \qquad \qquad \qquad\quad \textsuperscript{2} Microsoft \\
\hspace{-1.5em}{\tt\small \{shuquanye2-c,jingliao\}@cityu.edu.hk  \qquad \{yujiaxie,dochen,Yichong.Xu,luyuan,chezhu\}@microsoft.com}}

\maketitle

\begin{abstract}

This paper focuses on analyzing and improving the commonsense ability of recent popular vision-language (VL) models. Despite the great success, we observe that existing VL-models still lack commonsense knowledge/reasoning ability (e.g., ``Lemons are sour"), which is a vital component towards artificial general intelligence. Through our analysis, we find one important reason is that existing large-scale VL datasets do not contain much commonsense knowledge, which motivates us to improve the commonsense of VL-models from the data perspective. Rather than collecting a new VL training dataset, we propose a more scalable strategy, i.e., ``Data Augmentation with kNowledge graph linearization for CommonsensE capability'' (DANCE). It can be viewed as one type of data augmentation technique, which can inject commonsense knowledge into existing VL datasets on the fly during training. More specifically, we leverage the commonsense knowledge graph (e.g., ConceptNet) and create variants of text description in VL datasets via bidirectional sub-graph sequentialization.
For better commonsense evaluation, we further propose the first retrieval-based commonsense diagnostic benchmark. By conducting extensive experiments on some representative VL-models, we demonstrate that our DANCE technique is able to significantly improve the commonsense ability while maintaining the performance on vanilla retrieval tasks. The code and data are available at \textcolor{blue}{\url{https://github.com/pleaseconnectwifi/DANCE}}. %

\end{abstract}

\section{Introduction}
\label{sec:intro}

\begin{figure}[t]
  \centering
   \includegraphics[width=0.85\linewidth]{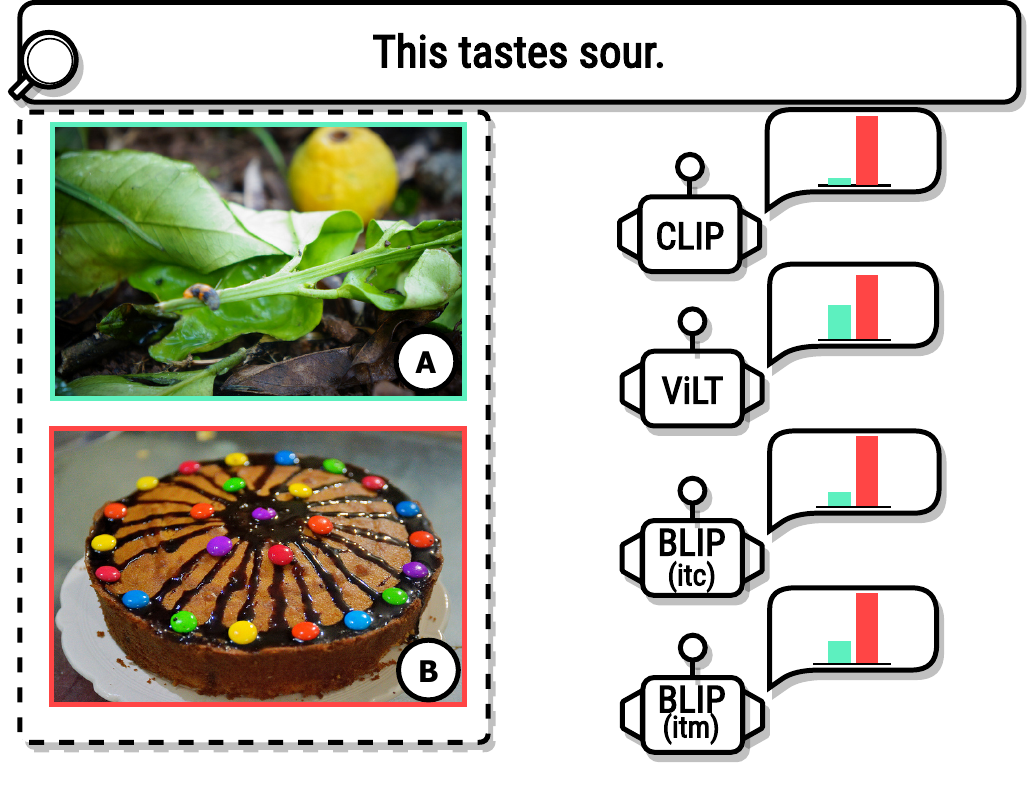}
\vspace{-1em}
   \caption{Illustration of the commonsense lacking problem of various popular VL-models, including CLIP~\cite{radford2021learning} pre-trained with contrastive supervision, ViLT~\cite{kim2021vilt} with matching supervision, and BLIP~\cite{li2022blip} with the both. The bar plots suggest the alignment scores of the images to the text. All models fail in retrieving the correct image with lemon (in blue).}\label{fig:teaser}
   \vspace{-1.5em}
\end{figure}

Many vision-based problems in our daily life go beyond perception and recognition. For example, when we hear people say ``\textit{It tastes sour}'', we need to identify they are talking about lemons on the table instead of the chocolate cake. Therefore, it is essential for artificial general intelligence to develop commonsense capability.    
Vision-Language (VL) models \cite{radford2021learning} recently show promising signals on mimicking the core cognitive activities of humans by understanding the visual and textual information in the same latent space \cite{zeng2022socratic}. However, we observed that VL-models, e.g., CLIP~\cite{radford2021learning}, still struggle when minor commonsense knowledge is needed. For example, as shown in figure \ref{fig:teaser}, none of the existing models correctly identify the lemon with text input ``\textit{It tastes sour}''. 

In this work, we take a step towards injecting the VL-models with commonsense capability. More specifically, we find one important reason for the commonsense lacking issue is that existing large-scale VL datasets do not contain much commonsense knowledge. On the one hand, regular VL datasets, e.g., COCO~\cite{lin2014microsoft} and CC 12M~\cite{changpinyo2021conceptual} contain more nouns and descriptive adjectives, with much fewer verbs and particles compared to regular texts. This distribution difference suggests that it might be infeasible for VL-models to gain commonsense capability by purely enlarging the data size, unlike language-only models~\cite{petroni2019language,jiang2020can}. Also, other objectives like visual question answering or generation are not widely applicable for training and have limited data size.

Inspired by the aforementioned findings, we propose Data Augmentation with kNowledge graph linearization for CommonsensE capability (DANCE). 
The main idea is to generate commonsense-augmented image-text pairs. 
To do so, one natural idea is to leverage the rich commonsense knowledge in knowledge graphs \cite{auer2007dbpedia,speer2017conceptnet}. However, it is not trivial to inject the knowledge into image-text pairs. On the one hand, structured data like graphs usually require specific architectures \cite{zhang2019graph,velivckovic2017graph} to embed, which is troublesome. On the other hand, if we associate the external knowledge with the text in the training stage, we will need the external knowledge-augmentation process in the inference stage as well to avoid domain shift \cite{shen2022k}. This is not desirable, since the corresponding knowledge is usually not available for the inference tasks. To address these challenges, we first re-organize the commonsense knowledge graph into entries with (entity, relation, entity) format, and pair them to the images that contain one of the entities. We then hide the name of entities in that image with demonstrative pronouns, e.g., ``\textit{this item}''. The generated descriptions are in textual form and therefore readily applicable for the training of most VL-models. More importantly, by forcing the model to memorize the relationships between entities in the training stage, such data augmentation is not needed in the inference stage. 
The data pair generation pipeline is automatic, scalable, and trustworthy, leveraging the existing consolidated commonsense knowledge base and the large and various collections of image-language supervision.  

In addition, existing VL commonsense evaluations are restricted to visual question answering and generation which are not a good fit or well received in the majority of VL-models. Therefore, we propose a new diagnostic test set in a wider adaptable form, i.e., Image-Text and Text-Image Retrieval, to achieve a fair evaluation of the pre-trained VL-models. The set is upgraded by neighborhood hard-negative filtering
to further ensure data quality. 

The effectiveness of the proposed strategy is validated by not only our diagnostic test set, but also the most generally used visual question answering benchmark for commonsense \cite{okvqa}.
Furthermore, we show that the commonsense capability of the models trained with DANCE can even generalize to unseen knowledge. 
We show the potential of the new train strategy and the test dataset via a deep study of its contents and baseline performance measurements across a variety of cutting-edge VL-models. Our main findings and contributions are summarized as follows:

\begin{enumerate}
    \item We propose a novel commonsense-aware training strategy DANCE, which is compatible with the most of VL-models. The inference stage needs no change. 
    \item We propose a new retrieval-based well-received commonsense benchmark to analyze a suite of VL-models and discover weaknesses that are not widely known: commonsense easy for humans (83\%) is hard for current state-of-the-art VL-models  ($<$42\%).  
    \item We conduct extensive experiments to demonstrate the effectiveness of the proposed strategy and diagnostic test set. The datasets and all the code will be made publicly available.
\end{enumerate}

\section{Related Work}
\label{sec:related}

\noindent\textbf{Vision-Language Contrastive Learning and Matching.} %
\label{sec:VLCVLM}
Vision-Language Contrastive Learning (VLC) and Matching (VLM), both of which aim to align vision and language, has been the fundamental tasks for Vision-Language model pre-training. They are the most commonly used
objectives~\cite{chen2022vlp}, and are used in the well-known
foundation models~\cite{wang2022omnivl,dong2022maskclip}: to name a few, CLIP~\cite{radford2021learning}, OwlVit~\cite{minderer2022simple}, ALIGN~\cite{jia2021scaling}, MDETR~\cite{kamath2021mdetr}, Florence~\cite{yuan2021florence} with VLC supervision; ViLT~\cite{kim2021vilt}, FLAVA~\cite{singh2022flava}, ViLBERT~\cite{lu2019vilbert}, UNITER~\cite{chen2020uniter}, Unicoder~\cite{li2020unicoder}, VisualBERT~\cite{li2019visualbert} utilize the VLM target; BLIP~\cite{li2022blip}, ALBEF~\cite{li2021align} uses them both. Many popular and large-scale image-text paired datasets~\cite{NIPS2011_5dd9db5e,young2014image,lin2014microsoft,krishna2017visual,sharma2018conceptual,changpinyo2021conceptual,schuhmann2021laion,schuhmann2022laion,miech19howto100m,Matterport3D,ye20213d,rostamzadeh2018fashion} are proposed on this task suitable for most common scenarios. Some of them target at specific cases like instructional video~\cite{miech19howto100m}, 3D scene~\cite{Matterport3D,ye20213d}, and fashion~\cite{rostamzadeh2018fashion}. However, most of them are not targeted at commonsense knowledge or reasoning.

\noindent\textbf{Image Text Retrieval.} %
Image Text Retrieval (ITR) is a typical cross-modal downstream task needing retrieving an image that matches a description most and vice versa. 
ITR can be naturally performed by the VL-models pre-trained on VLC/VLM targets, and widely received in the majority of the VL-models~\cite{ijcai2022p0759} even in a zero-shot manner. Though, how to improve the commonsense knowledge of ITR still requires further study.
CVSE~\cite{Wang2020CVSE} injects commonsense knowledge into VL-models for ITR using statistical correlations in caption corpus. However, such knowledge is constrained by the correlations of corpus and is not a perfect fit for ITR~\cite{ijcai2022p0759}.
Various datasets~\cite{lin2014microsoft,young2014image,wu2021fashion,liu2021image,olondriz2021foodi} have been proposed to evaluate the ITR system. However, most of them do not question VL-models' commonsense ability.

\noindent\textbf{Commonsense in Vision-Language Datasets.}
Several studies on commonsense over visual-language input can be divided into two branches according to the task type.
The first branch includes Visual Question Answering (VQA) datasets~\cite{wang2017explicit,fvqatpami,jain2021select,okvqa,ye20213d}. The model is required to give a natural language answer to a specific question-image pair, for which commonsense knowledge is required~\cite{lin2022revive}. 
However, performing VQA evaluation automatically is not trivial for most of the VL-models, especially for dual-encoder architecture, e.g., CLIP, as studied in~\cite{du2022survey}.
Besides, as their data collection requires plenty of human effort, their image amount and language corpus of are quite small.
Another branch focuses on generation or multi-choice tasks. Visual Commonsense Reasoning~\cite{zellers2019vcr} evaluates the commonsense ability via two multi-choice tasks: question answering and answer justification. VisualCOMET~\cite{park2020visualcomet} evaluates the commonsense ability via inferencing events and intents. However, they are collected from movie clips, and the commonsense knowledge required is focused on humans and events, which limits the variety of image and knowledge. Also, transferring from commonly used pre-training tasks to generation or multi-choice tasks itself is a challenging task~\cite{li2021supervision,mu2021slip,yao2021filip}.
To evaluate and improve the commonsense ability of VL-models
, we take the first step towards automatic and direct applicable commonsense evaluation via ITR, along with a scalable learning strategy suitable for VLC/VLM with variable augmented commonsense.

\noindent\textbf{Commonsense in NLP.} %
Commonsense and knowledge representation has a long-period development in NLP~\cite{zhu2022knowledge,xu2021human,xu2020fusing}, with plenty of famous knowledge bases~\cite{lenat1993building,singh2002public,anacleto2006can,auer2007dbpedia,singhal2012introducing,speer2013conceptnet,speer2017conceptnet} emerged. ConceptNet~\cite{speer2017conceptnet} is a popular and consolidated commonsense knowledge graph with 8 million nodes and 21 million edges collected from various sources including expert-created, crowd-sourcing, and games. 
While there are early explorations on general VL-model training methods with external knowledge \cite{shen2022k}, we are the first one that aims to train a general-purpose commonsense-aware VL-model, with no restrictions in the inference stage.

\section{Data Augmentation Strategy with Knowledge Graph Linearization}
We present our DANCE training strategy that enhances the commonsense ability of VL-models via learning with novel and scalable commonsense augmented data.

In short, we augment existing image-text pairs with knowledge graphs. Denote the set of image-text pairs as $\mathcal{D}=\{(i_k, t_k)\}_{k=1}^K$, where $i_k$ and $t_k$ are paired images and texts. Denote the knowledge graph as $\mathcal{G}=(\mathcal{V}, \mathcal{E})$, where $\mathcal{V}$ is the node set and $\mathcal{E}$ is the edge set. Each edge $e\in\mathcal{E}$ is a tuple $(v_i, r, w, v_j)$, with $v_i, v_j \in \mathcal{V}$, $r$ is the commonsense relationship pointing directionally from $v_i$ to $v_j$, and $w$ is the weight highlighting the importance of this edge. Here, $v_i$ is denoted as the \textit{head}, while $v_j$ is the \textit{tail}.
For example, a directed edge from ConceptNet \cite{speer2017conceptnet} takes the form as 
\begin{equation*}
    (\text{``\textit{Net}''}, \text{``\textit{is used for}''}, 0.3,  \text{``\textit{catching fish}''}).
\end{equation*}
\begin{figure*}[t]
  \centering
   \includegraphics[width=0.8\linewidth]{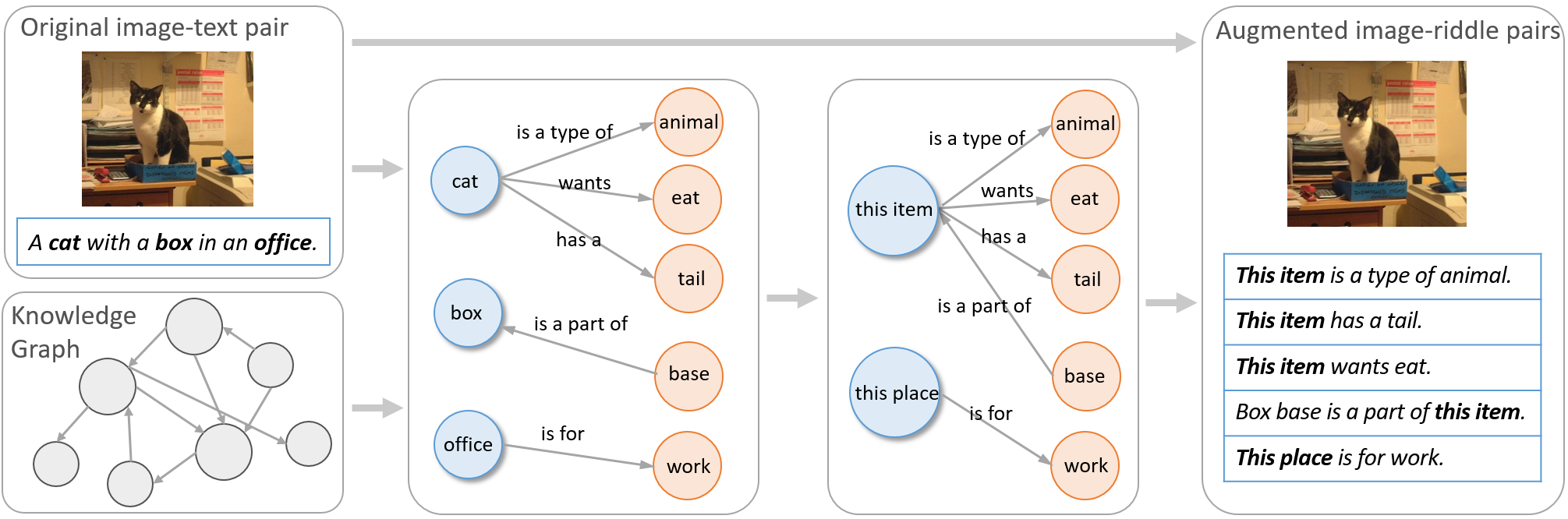}
\vspace{-1em}
   \caption{Illustration of VLCR construction process of DANCE.  }\label{fig:VLCRillust}
   \vspace{-1.2em}
\end{figure*}
We explore an automatic way to generate paired commonsense-aware training data by augmenting commonsense knowledge into VL datasets.
The automatic data construction process is shown in~\Fref{fig:VLCRillust}.

\subsection{Extraction of Image-Entity Pair} 
To pair an image $i_k$ with its corresponding knowledge, we first need to find out what entities are in it. Given the presence of the corresponding descriptive text $t_k$, a reliable way is to extract the linguistic entities from $t_k$  via natural language processing toolkits\footnote{For example, NLTK \cite{bird2009natural} and SpaCy \cite{spacy2}, two popular libraries for natural language processing in Python.}. For example, for the image in the upper left of \Fref{fig:VLCRillust}, the corresponding caption is ``\textit{A cat with a box in an office}''. Using standard toolkits, we can obtain the entities as ``\textit{cat}'', ``\textit{box}'', and ``\textit{office}''. 
In detail, we extract the linguistic entities $\mathcal{N}_k = \{\epsilon_{k,i}\}_{i=1}^{I}$, and further perform format cleaning that removes determiners or adjectives, and filtering out entities that are too general to be meaningful to get a subset $\hat{\mathcal{N}}_k\subseteq\mathcal{N}_k$. In this way, we obtain the entities corresponding to image $i_k$.

\subsection{Bidirectional Sub-Graph Sequentialization} 
With the list of entities in each image, we perform bidirectional sub-graph sequentialization to obtain a list of commonsense riddles in textual format, which can be readily used for contrastive training. The key idea is to find commonsense knowledge descriptions associated with each entity in the image, but with the entities' names hidden by demonstrative pronouns. The pipeline of our graph operation can be summarized as follows:
\begin{enumerate}
    \item To collect the commonsense knowledge associated with image $i_k$, we first query the directed knowledge graph to obtain sub-graphs where the nodes are connected, in either direction, to at least one entity in $\hat{\mathcal{N}}_k$.
    \item Hide the names of the entities in the image by replacing the different subject nodes with ``\textit{this}'' nodes.
    \item We perform sequentialization to translate the sub-graph into a list of subject-hidden commonsense knowledge descriptions in textual format.
\end{enumerate}

$\bullet$ \fontsize{10}{11}{\textbf{Bidirectional sub-graph query.}} Specifically,  we query the sub-graph that relates to $i_k$ from the directed graph $G$, so that each edge within it is connected with at least a node representing one entity in $\hat{\mathcal{N}}_k$. The connection we need to check is bidirectional: both head and tail should be taken into account. Formally, we perform a bidirectional query to get sub-graphs,
\begin{align*}
 G_k = &(V_k, E_k) \\
\text{s.t.  } V_k =& \left\{v | v \in \ell_\mathcal{E}(u), u\in \hat{\mathcal{N}}_k \right\} \cup \hat{\mathcal{N}}_k,\\
  E_k  = & \left\{(u, r, w, v)\in \mathcal{E}| u\in \hat{\mathcal{N}}_k \right\} \\
  & \cup \left\{(v, r, w, u)\in \mathcal{E}| u\in \hat{\mathcal{N}}_k \right\},
\end{align*}
and $\ell_\mathcal{E}(u)$ is the neighbors of node $u$ when the edge set is $\mathcal{E}$. 
We end up with a sub-graph $G_k$. We refer the nodes that are directly from $\hat{\mathcal{N}}_k$ as subject nodes, i.e., $u \in \hat{\mathcal{N}}_k$.

$\bullet$ \fontsize{10}{11}{\textbf{Hiding subject names via node substitution.}}
After querying the sub-graphs, we perform node substitution that replaces subject nodes with ``\textit{this}'' nodes, to hide the names of all the entities in the image. 
Specifically, we construct a mapping
\begin{equation*}
f(\cdot)\colon \mathcal{V} \to \mathcal{S},
\end{equation*}
that maps the actual entity nodes to a substitution set $\mathcal{S}$, which is defined as
\begin{equation*}
    \mathcal{S} = \left\{ \text{``\textit{this item}''}, \text{``\textit{this person}''}, \text{``\textit{this place}''} \right\}.
\end{equation*}
In detail, a node $u$ is mapped to ``\textit{this person}'' if it belongs to a ``\textit{person}'' word, e.g., ``\textit{lady}'', ``\textit{guy}'', or mapped to ``\textit{this place}'' if it has location property\footnote{For example, in ConceptNet, such nodes has relation property ``\textit{AtLocation}''.} or the subject name matches Places356 categories~\cite{zhou2017places}, or ``\textit{this place}'' if neither of above is matched. 
Further, we filter out some edges in the sub-graph with weights below a certain threshold $\tau$. More rigorously, the graph after substitution is
\small
\begin{align*}
G'_k  = & \left(V'_k, E'_k\right),  \\
\text{s.t.  } E'_k = & \left\{(f(u), r, w, v)| (u, r, w, v)\in E_k, u\in \hat{\mathcal{N}}_k, w>\tau \right\}  \\
   \quad \quad & \cup \left\{(v, r, w, f(u))|(v, r, w, u)\in E_k, u\in \hat{\mathcal{N}}_k , w>\tau\right\}, \\
 V'_k  = & \left\{v | v \in \ell_{E'_k}(u), u\in \hat{\mathcal{N}}_k \} \cup \{f(u) |u\in \hat{\mathcal{N}}_k\right\}.
\end{align*}
\normalsize

$\bullet$ \fontsize{10}{11}{\textbf{Riddle generation.}}
We concatenate the head, the relation, and the tail in each edge to generate riddles in natural language. Take an example from \Fref{fig:VLCRillust}. For edge $(\text{``\textit{this item}''}, \text{``\textit{is a type of}''}, 0.6, \text{``\textit{animal}''} ),$
we re-format it into natural language riddle $\text{``\textit{this item is a type of animal}''}$
and pair it back to image $i_k$. 
In this way, we obtain multiple image-riddle pairs for each image, which is readily usable for contrastive VL-model training. 
The whole process is performed automatically without human involvement. Leveraging mature image-text datasets and knowledge graphs, the quality of the generated data is well guaranteed.

\subsection{Training Strategy}
Since image-riddle pairs are essentially image-text pairs, by mixing them with the existing VL databases with a certain ratio, we can pre-train or fine-tune the VL-model without changing the model architecture and other training settings. However, since the total amount of text generated by DANCE is several times larger than that of the existing VL dataset, simply merging our data with the original dataset will cause our data to dominate. Denote the proportion of the augmented data in the training batch as $p$. We observe that a larger $p$ at the beginning and a smaller $p$ in the later training stage can lead to good performance. Therefore, we adopt a curriculum learning strategy, with linearly decreasing $p$ \cite{soviany2022curriculum}.  
In this way, the percentage of original and our data sources can be controlled dynamically. 
There is no change to the inference stage.

\section{Diagnostic Data and Automatic Evaluation}

It is still an open problem to automatically and directly compare commonsense knowledge of VL-models without transferring to other downstream
tasks.
Thus, we introduce a diagnostic test set for comparison of the commonsense ability, in the form of a retrieval task compatible with various VL-models. Our task is divided into Text-Image and Image-Text retrieval. The former one is to retrieve which image best matches the description that requires commonsense but with the referred subject hidden, and the latter is vice versa. Thus, we mainly focus on the former one in the following paragraph of generation and evaluation. Formally, the model is either given a riddle $d$ with a list of $N_\text{i}$ candidate images $\mathcal{I}=\{i_1,...,i_{N_\text{i}}\}$ to choose from, or an image with a list of riddles.
Models need to score the alignment between the images and the riddles and return a sorting of them. The data construction is based on the COCO test set and ConceptNet.

\noindent\textbf{Generation of candidate set.} %
Different from existing image-text datasets that usually contain one-to-one pairs, in the generated text set, there are multiple positive riddles for each image, and multiple positive images for each riddle. 
Thus, for test data, we also need to generate candidate sets including both positive and negative images with a consistent ratio. Suppose in the image list $\mathcal{I}$, there are $n$ positive images $i_1,...,i_n$. They are chosen at random from the set of images that all contains the substituted entity in the riddle $d$. To ensure the high quality of the negative images, i.e., $i_{n+1},...,i_{N_{\text{i}}}$, rather than random sampling from all possible negatives, we design to search for hard-negative samples. Specifically, We employ neighborhood hard-negative filtering, i.e., we find images whose entities are highly related to the subject entity, but none of these entities satisfy the riddle. To capture the correlation between entities, we use the graph distance in ConceptNet, i.e., from one entity, we filter for their nearest neighbor entities that are connected by the edges among three relationships: ``\textit{RelatedTo}'', ``\textit{DistinctFrom}'', and ``\textit{Antonym}''. We construct the image-to-riddle data in the same way. 
 
\noindent\textbf{Generalization ability.} %
To further diagnose the ability to infer new knowledge by using existing commonsense knowledge, we randomly hold out some knowledge from training. For example, given that ``\textit{pineapple can be found on a pizza}'', and ``\textit{pizza hut is one of the makers of pizza}'', we want to see whether the model can reason that ``\textit{pineapple may be required by pizza hut}''. 
Therefore, we further divide the test set into two splits: \textbf{test-seen} split, in which all knowledge appears in the training set, and \textbf{test-unseen} split where the corresponding relationships are not present in training, to see whether the model can reason about new knowledge using the existing ones.
We also enforce that all the images in the two test splits are not present in training.

\noindent\textbf{Automatic evaluation protocol.} %
For automatic evaluation, we adopt perplexity score as the evaluation metric, following the works~\cite{das2017visual, park2020visualcomet}. In the experiment, we set the candidate number for each sample as $50$, with the number of positive samples $n$ between $1$ to $15$, and measure the average accuracy of retrieved ground truth inference, denoted as $Acc@50$. Our evaluation protocol is based on retrieval tasks, and therefore is compatible with most of the VL-model architectures.  

\begin{figure}[t]
  \centering
   \includegraphics[width=1.0\linewidth]{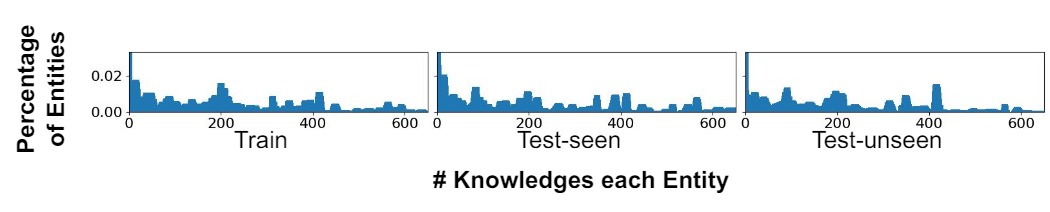}
\vspace{-2.2em}
   \caption{Visualization of the distributions of our train and Test-Image test splits.}\label{fig:DataDistributionTestUnseen}
   \vspace{-1.5em}
\end{figure}

\noindent\textbf{Distribution of test-unseen split.} We visualize the distribution of test-unseen split in comparison with training and test-seen split to verify that it represents a reasonable knowledge distribution. In \Fref{fig:DataDistributionTestUnseen}, we show the distributions of the train, test-seen, and test-unseen based on COCO and ConceptNet. The $x$ axis represents the number of commonsense knowledge descriptions associated with each entity, and the $y$ axis represents the percentage of these entities among all entities. The distribution of the test-unseen split does not shift much from the training and test-seen splits.

\section{Experiment}

\begin{table}[t]
\setlength{\tabcolsep}{0.3mm}
\centering
\scalebox{0.8}{
\begin{tabular}{@{}cc|cccc@{}}
\toprule
\multicolumn{2}{c|}{\multirow{2}{*}{Models}} & \multicolumn{2}{c}{Text-Image} & \multicolumn{2}{c}{Image-Text} \\
\multicolumn{2}{c|}{}                        & test-seen↑    & test-unseen↑   & test-seen↑    & test-unseen↑   \\ \midrule
Random         &  -                           &      0.2381&	0.2380&	0.2400&	0.2362
              \\ \midrule
Contr-    & CLIP(ViT-L)        &      0.3951&	0.3949&	0.3817&	0.3961
                \\
astive        & OwlVit(ViT-L)            &      0.3673&	0.3644&	0.3325&	0.3230
               \\ \midrule
Matching       & ViLT(ViLT-B)   &      0.4098&	0.4077&	0.3217&	0.3534
               \\
               & FLAVA(ViT-B)             &      0.4144&	0.4093&	0.3850&	0.3843
                \\ \midrule
Both           & BLIP(ViT-L itm)             &       0.4030&	0.4019&	0.4017&	0.4194
              \\
               & BLIP(ViT-L itc)             &       0.3835&	0.4007&	0.3167&	0.3100
               \\
               & ALBEF(ViT-B)                &      0.3901&	0.3792&	0.3749&	0.3832
              \\ \midrule
Human          & -                           &      0.8202&	0.8023&	0.8497&	0.8521
              \\ \bottomrule
\end{tabular}}
\vspace{-0.8em}
\caption{Comparison with various state-of-the-art VL-models and human performance on our diagnostic test set.}
\vspace{-0.4em}
  \label{tab:vshuman}
\end{table}

In this section, we first highlight the commonsense lacking issue in both the popular VL datasets and the existing VL-models, then provide more analysis on the augmented training data, and finally provide detailed empirical evidence showing the effectiveness of the proposed DANCE strategy.

\begin{figure}[t]
  \centering
   \includegraphics[width=0.95\linewidth]{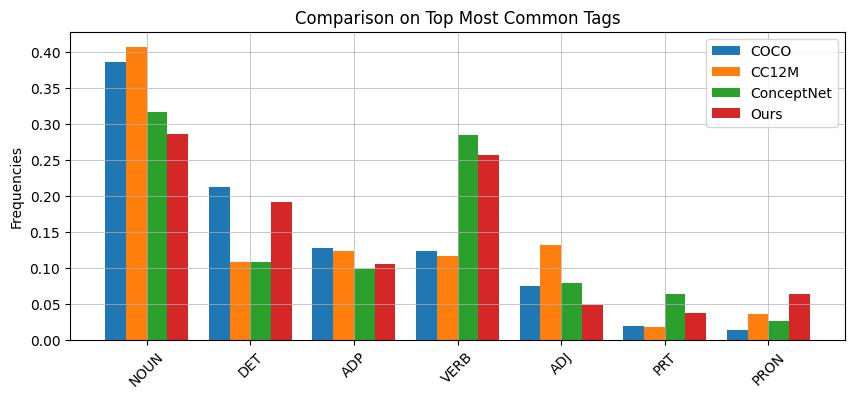} \\
   \vspace{-0.4em}
   \includegraphics[width=0.95\linewidth]{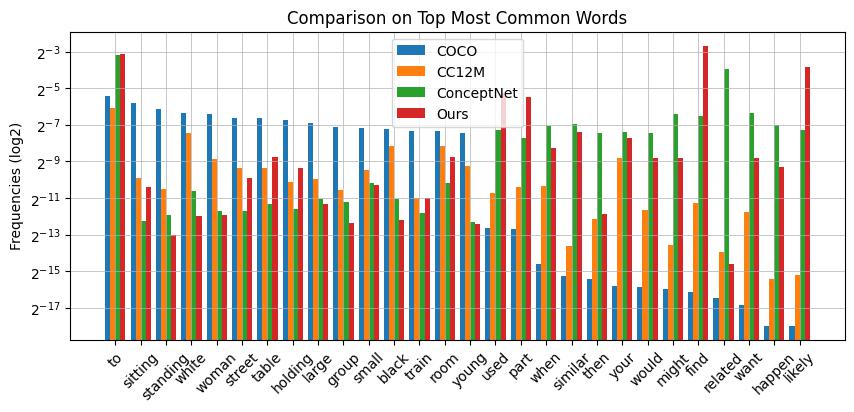}
\vspace{-1.5em}
   \caption{Comparison of the syntactic categories and words distributions of fundamental VL data (COCO~\cite{lin2014microsoft} and CC12M~\cite{changpinyo2021conceptual}), NLP knowledge base (ConceptNet~\cite{speer2017conceptnet}) and ours. Commonsense is lacking in VL data, but it has significantly improved in ours.}\label{fig:DistributionCOCOvsOurs}
   \vspace{-1.1em}
\end{figure}

\subsection{Commonsense Lacking Issue}

\noindent\textbf{Commonsense lacking in fundamental VL data.}
We find that the current fundamental VL datasets, on which VL-models are trained, are to blame for the commonsense knowledge-lacking issue.
We show that the current fundamental datasets for VL-models can not provide sufficient commonsense knowledge compared to regular texts.

We illustrate the issue by comparing the most popular VL datasets (COCO~\cite{lin2014microsoft} and CC 12M~\cite{changpinyo2021conceptual}) with the commonly used language-only knowledge bases (ConceptNet~\cite{speer2017conceptnet}) in terms of the distributions of the syntactic categories and words.
In the upper part of \Fref{fig:DistributionCOCOvsOurs}, we compare the distributions of the most frequent part-of-speech (POS) tags with punctuation marks excluded. In the lower part, we show the comparison of the most frequent word tokens. There is a significant difference between top POS tag / word token distributions of VL datasets compared with those of the regular texts. We note that most frequent words in the text in existing VL datasets are nouns, especially for individual entities that appear in the images and the corresponding caption (e.g., ``\textit{woman}'', ``\textit{train}'', ``\textit{room}''). In contrast, the knowledge base ConceptNet has comparatively many more verbs, e.g., ``\textit{used}'', ``\textit{would}'', ``\textit{find}'', ``\textit{want}'', ``\textit{happen}'', ``\textit{requires}'', which contain richer information about the relationship between entities. In addition, the knowledge base includes more particles (PRT), like ``\textit{to}'', and pronouns (PRON) like ``\textit{your}'', which are associated with interconnection information. 

In order to develop common sense and reasoning ability, in addition to knowing each isolated entity, there is a high demand for rich but implicit information about the relationships and interconnections between entities.  Thus, the fundamental VL dataset, which is primarily occupied by information about individual entities that appear explicitly, does not meet the requirements of VL-models for common knowledge, in terms of both learning or evaluation. This implies that we should enhance VL data with commonsense. Our augmented data via DANCE provides significantly more commonsense knowledge than the VL datasets. In Sec. 1 of the appendix, we include additional comparisons of VL data with common NLP data to further illustrate the commonsense lacking issue.

\noindent\textbf{Baselines vs Human.}
Here we show the performance of various state-of-the-art VL-models on our diagnostic test set. Specifically, we consider VL-models in three categories: models trained by contrastive supervision, e.g., CLIP~\cite{radford2021learning}, OwlVit~\cite{minderer2022simple}, by matching supervision like ViLT~\cite{kim2021vilt}, FLAVA~\cite{singh2022flava}, and by the both, such as BLIP~\cite{li2022blip}, ALBEF~\cite{li2021align}. It is either impossible or difficult to directly test these models against knowledge-based benchmarks designed with other downstream tasks.
As an additional reference for performance comparison, we also report the performance of random ordering as the lower bound, and human performance on a random sub-set with $50$ samples each split as the upper bound. 
All the mentioned models are with their official checkpoints. CLIP is with ViT-L/14 backbone pre-trained on 400M images at 336-pixel resolution, ViLT is with ViLT-B/32 backbone pre-trained on 21M images and fine-tuned on COCO retrieval, and BLIP is with ViT-L/16 backbone pre-trained on 129M images with bootstrapping and find-tuned on COCO retrieval.

Results with our automatic evaluation metric $Acc@50$ are shown in~\Tref{tab:vshuman}. Through observation, we discover that, while most of the retrieval is easy for humans (83\% on average, 81\% for Text-Image, and 85\% for Image-Text), they are hard for current state-of-the-art VL-models  ($<$40\% on average, 39.4\% for Text-Image, and 36.3\% for Image-Text) whose performances are only slightly better than the random result (24\% on average).

\subsection{Analysis on the Augmented Data}
\noindent{\fontsize{10.5}{12}{\textbf{Implementation Details.}}}
Before the bidirectional query, we need to match the natural language words in text $t_k$ to the knowledge graph entities. Therefore, we perform  Unicode normalization, lower-casing, punctuation cleaning, underscoring and pre-pending to the words. For example, the English words ``\textit{A traffic jam}'' becomes ``\textit{/c/en/traffic\_jam}'' by the standardization, so that it is matched to an entity. The threshold $\tau$ is set to 0.5. 
We are based on two mature human-annotated image-text paired datasets COCO~\cite{lin2014microsoft}, VG~\cite{krishna2017visual}, and three web datasets SBU captions~\cite{ordonez2011im2text}, Conceptual Captions (CC3M)~\cite{sharma2018conceptual}, and Conceptual 12M (CC12M)~\cite{changpinyo2021conceptual}, with 14M images in total.
Our image splitting for COCO follows a popular split~\cite{anderson2018bottom,rohrbach2018object,yang2019auto,wang2020show,li2022blip,wang2022omnivl,hao2022language,dai2022enabling,tewel2021zero}.

\noindent{\fontsize{10.5}{12}{\textbf{Dataset comparison with various knowledge-based datasets.}}}
\begin{table}[t]
\setlength{\tabcolsep}{0.3mm}
\centering
\scalebox{0.73}{
\begin{tabular}{@{}c|ccccc@{}}
\toprule
               & DANCE              & VCR            & Visual-  & OK-VQA  & S3VQA   \\
               & (this work)       &\cite{zellers2019vcr} & COMET~\cite{park2020visualcomet}  & \cite{okvqa} & \cite{jain2021select}        \\ \midrule
supervision    & contrastive/matching         & multi-choice   & inference    & VQA      & VQA      \\
\# images      & 14.1M+            & 0.1M           & 59K          & 14K     & 7K      \\
\# texts       & 447M+             & 0.3M           & 1.5M         & 14K     & 7K      \\
knowledge & general & people action & movie event & factoid & factoid \\ \bottomrule
\end{tabular}}
\vspace{-1em}
  \caption{Comparison with various knowledge-based datasets. }
  \vspace{-1.2em}
  \label{tab:CompareVQADatasets}
\end{table}
In \Tref{tab:CompareVQADatasets}, we compare the training set generated by DANCE to relevant knowledge-based datasets and display their properties and statistics. We are the first commonsense knowledge-based dataset to focus on contrastive or matching supervision. We have larger-scale multi-source images and a corpus compared to the relevant datasets which are challenging to gather at scale, and we can expand even further if other image-text datasets or knowledge bases are included in the generation process. Also, our dataset includes various general knowledge from a consolidated commonsense knowledge graph, while the knowledge type in some of the relevant datasets (e.g., VCR, VisualCOMET) is limited to people or events in films.

\begin{figure}[t]
  \centering
   \includegraphics[width=0.42\linewidth]{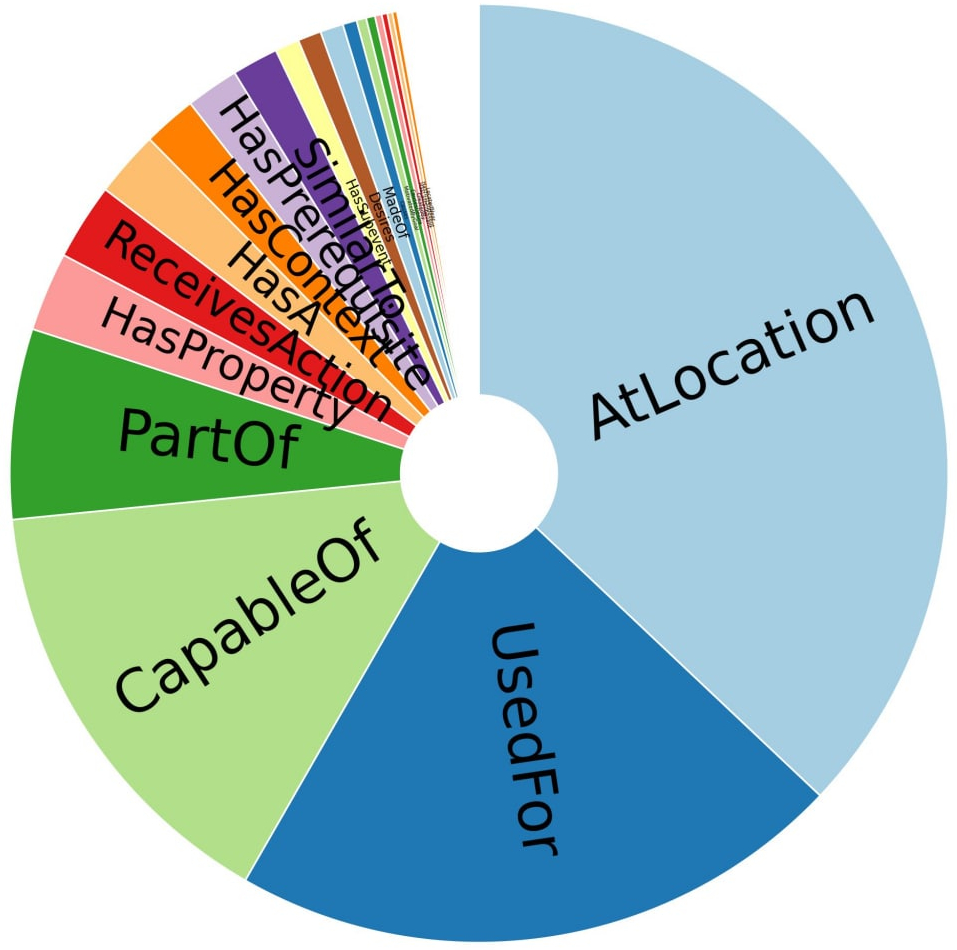}
   \includegraphics[width=0.39\linewidth]{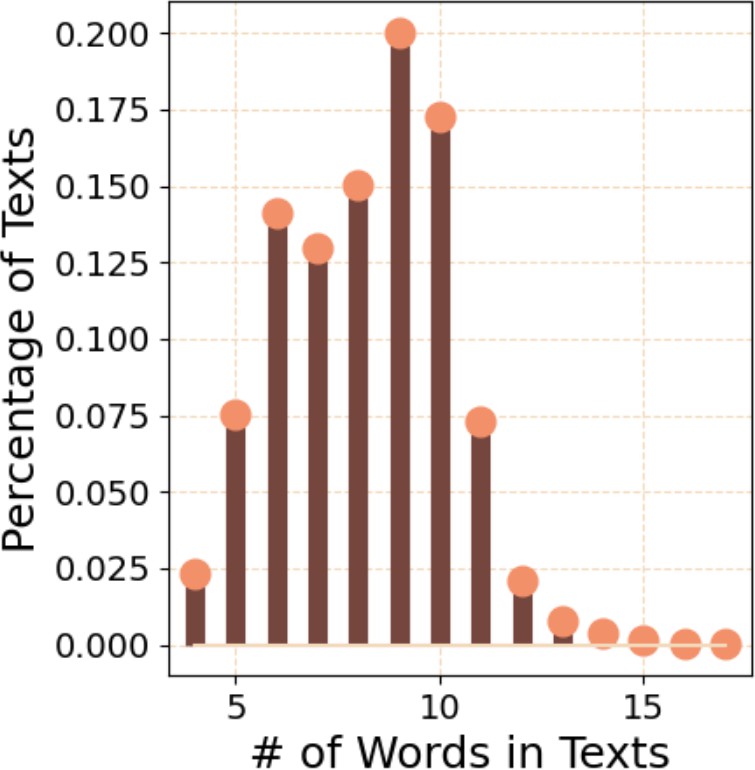}
\vspace{-1em}
   \caption{Left: Commonsense knowledge type distribution. Right: Length distribution of each commonsense riddle.}\label{fig:KnowledgeDistributionTrain}
   \vspace{-1em}
\end{figure}

\noindent{\fontsize{10.5}{12}{\textbf{Other statistics.}}}
The left part of \Fref{fig:KnowledgeDistributionTrain} shows the distribution of commonsense knowledge type in the training set. We can observe that various types of commonsense knowledge are included in our generated data.
The right part is the distribution of the average length of the commonsense riddles in our generated training set. The average riddle length generated from COCO and ConceptNet is $8.19$.%

\subsection{Effectiveness of DANCE}

\noindent{\fontsize{10.5}{12}{\textbf{Training details.}}}
All pre-training is done on four nodes, each with eight A100 GPUs. We use ViT pre-trained on ImageNet~\cite{dosovitskiy2020image} to initialize the image transformers, and BERT-base~\cite{devlin2018bert} for the text transformers. Two backbones are used: ViT-Base/16 and ViT-Large/16. We pre-train the model of ViT-B backbone with batch size $2,800$, and ViT-L backbone with batch size $2,400$, for total 20 epochs. We use the AdamW optimizer, training with learning rate warm-up to $3\times 10^{-4}$ for ViT-B and $2\times 10^{-4}$ for ViT-L, and decaying linearly with the rate $0.85$. For fine-tuning on retrieval tasks, the learning rate is warm-up to $1\times 10^{-5}$ for ViT-B and $5\times 10^{-6}$ for ViT-L with $6$ epoch. For fine-tuning on question-answering tasks, the learning rate is warm-up to $2\times 10^{-5}$ with a total $10$ epoch. We use the same supervision and model architecture as BLIP~\cite{li2022blip}. Pre-training data is the same as~\cite{li2021align,li2022blip}. It is partially or fully augmented by DANCE, ending up with $0.4$ billion and $40$ million image-text pairs respectively. Noted that despite the data scale being greatly improved by DANCE, in both pre-training and fine-tuning, all methods are trained with the same number of data batches and steps for fairness.

\noindent{\fontsize{10.5}{12}{\textbf{DANCE for Pre-training.}}}
In this section, we show the effect of the DANCE strategy when used only during the pre-training phase. To provide a comprehensive study, we compare the performance of various model architectures and training corpora with and without DANCE.
To ensure fairness, we adopt the data volumes to be equal, meaning that DANCE and the corresponding baseline are pre-trained and fine-tuned on the same number of data batches and iteration steps. Simultaneously, DANCE does not introduce more pre-training images than corresponding baseline. In other words, DANCE is entirely based on existing images.

$\bullet$ \fontsize{10}{11}{\textbf{Comparison on our diagnostic test set and vanilla retrieval benchmark.}}
We compare models with or without DANCE in the pre-training stage, while all models in the fine-tuning stage are trained only on the COCO retrieval dataset without DANCE.

\begin{table*}[t]
\setlength{\tabcolsep}{1.2mm}
\centering
\scalebox{0.87}{
\begin{tabular}{@{}ccc|cccccc@{}}
\toprule
\multirow{3}{*}{Backbone} & \multirow{3}{*}{Pre-train} & \multirow{3}{*}{Fine-tune} & \multicolumn{4}{c}{Ours test set}                               & \multicolumn{2}{c}{COCO (5K test set)}          \\
                          &                            &                            & \multicolumn{2}{c}{Text-Image} & \multicolumn{2}{c}{Image-Text} & \multirow{2}{*}{TR@1↑} & \multirow{2}{*}{IR@1↑} \\
                          &                            &                            & test-seen↑    & test-unseen↑   & test-seen↑    & test-unseen↑   &                        &                        \\ \midrule
\rowcolor{LightLightGray}
ViT-B                     & 14M                        & COCO                       &  0.3986&	0.3892&	0.2916&	0.3301&	78.40& 60.70                       \\
\rowcolor{LightGray}
ViT-L                     & 14M                        & COCO                       &   0.4030&	0.4019&	0.4017&	0.4194&	81.12&	63.96                     \\
\rowcolor{LightLightGray}
ViT-B                     & 14M+DANCE(part)             & COCO                       &  0.5107&	0.5141&	0.5252&	0.5053&	78.80&	60.48                      \\
\rowcolor{LightGray}
ViT-L                     & 14M+DANCE(part)             & COCO                       &  0.5408&	0.5326&	0.5363&	0.5113&	80.89&	64.15                     \\
\rowcolor{LightGray}
ViT-L                     & 14M+DANCE(whole)            & COCO                       &   \textbf{0.5721}&	\textbf{0.5458}&	\textbf{0.5600}&	\textbf{0.5242}&	\textbf{81.92}&	\textbf{65.26}                  \\ \midrule
\rowcolor{LightLightGray}
ViT-B                     & 14M                        & COCO+DANCE                  &   0.4566&	0.4077&	0.3421&	0.3845&	77.94&	60.83                    \\
\rowcolor{LightGray}
ViT-L                     & 14M                        & COCO+DANCE                  &   0.4610&	0.4333&	0.4565&	0.4395&	81.86&	64.17                    \\ \bottomrule
\end{tabular}}
\vspace{-0.9em}
\caption{Effect of DANCE for pre-training (first five rows) and fine-tuning (last two rows), testing on ours test set and COCO retrieval.}
\vspace{-1.3em}
\label{tab:VLCRpretrainfinetune}
\end{table*}

The results are shown in the first five rows of~\Tref{tab:VLCRpretrainfinetune}. In all pre-training setups, the DANCE pre-training consistently outperforms the corresponding baseline on our test set by a large scale. By augmenting a part of the pre-training data with DANCE, an average improvement of 16\% and 12\% can be seen on ViT-B and ViT-L. And by using DANCE on the entire pre-train data, we achieve 14\% improvements on average. In addition, on the unseen splits which contain commonsense knowledge held out from training data, significant improvements can also be observed. This shows that DANCE pre-training not only improves the commonsense ability of the model, but also empowers their ability to generalize to new knowledge on the basis of existing commonsense knowledge.
The performance is maintained or even better on the vanilla benchmark of COCO retrieval, which does not contain much commonsense. This demonstrates that DANCE enhances the commonsense ability and learns general visual-linguistic representations at the same time. 
Yet, there is still significant headroom in comparison with human scores (83\%).

\begin{figure}[t]
  \centering
   \includegraphics[width=\linewidth]{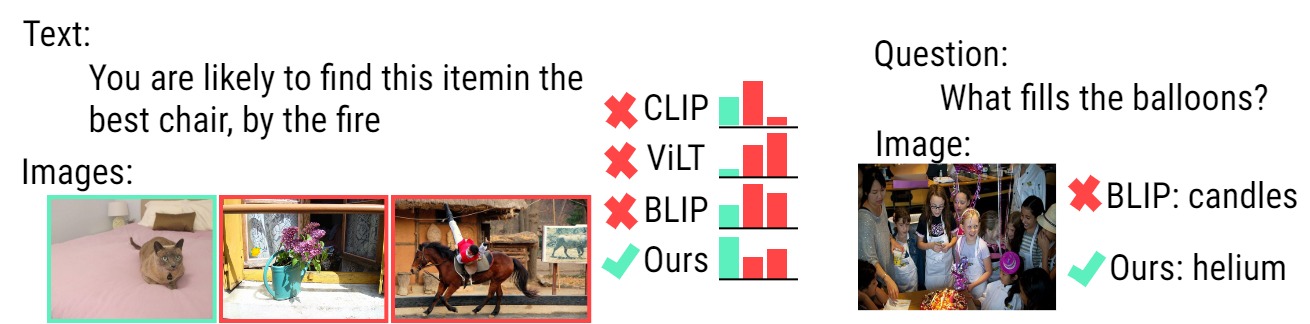}
\vspace{-1em}
   \caption{Qualitative examples from the diagnostic test set (left) and OK-VQA (right).}\label{fig:qualitative}
   \vspace{-1.1em}
\end{figure}

We also show the qualitative results of different VL-models in our diagnostic test set in~\Fref{fig:qualitative}. One the left is a text-image retrieval sample from the text-unseen split of our test set. The baseline model fails while our DANCE pre-trained model successfully retrieves the correct image, even though the knowledge is excluded from our training set. This also illustrates the reasoning ability enhanced by DANCE. On the right, with the OK-VQA question, the model pre-trained by DANCE correctly answers with helium. We own it to DANCE-augmented commonsense, from the knowledge base including the knowledge that helium is used for filling party balloons~\cite{Balloon}.
Overall, the above experiments demonstrate that pre-training with DANCE remarkably enhances the commonsense and even reasoning ability of VL-model. In Sec. 2 of the appendix, we include additional qualitative results.

$\bullet$ \fontsize{10}{11}{\textbf{Comparison on existing commonsense benchmark.}}
We also perform experiments on the commonly used commonsense knowledge benchmark. Here we choose the popular crowdsourced benchmark OK-VQA~\cite{okvqa} benchmark. In this benchmark, the commonsense ability of the model is evaluated via the downstream Visual Question Answering (VQA) task, which answers questions about an image that requires commonsense. It means that this task cannot be performed directly by the pre-trained baselines. Thus, to evaluate on this benchmark, we apply rearrangement of the model architecture during fine-tuning, following implementation by BLIP~\cite{li2022blip} for adopting to VQA~\cite{antol2015vqa}. We also note that the OK-VQA benchmark is crowdsourced and ``\textit{open}''-domain, which means that there is no data leakage in the commonsense knowledge graph used for training.

The baselines are BLIP
pre-trained on 14M images including COCO~\cite{lin2014microsoft}, VG~\cite{krishna2017visual}, SBU captions~\cite{ordonez2011im2text}, Conceptual Captions (CC3M)~\cite{sharma2018conceptual}, and Conceptual 12M (CC12M)~\cite{changpinyo2021conceptual}, and fine-tuned on OK-VQA.
They are compared with DANCE pre-trained models respectively. Performing DANCE on a part of the pre-train data (COCO) generating $40$M image-text pairs during the pre-training is denoted by DANCE(part), and on the full data generating $0.4$B image-text pairs is denoted by DANCE(whole).

\begin{table}[t]
\setlength{\tabcolsep}{0.8mm}
\centering
\scalebox{0.86}{
\begin{tabular}{@{}cccc@{}}
\toprule
Backbone & Pre-train       & Fine-tune & OK-VQA Acc↑ \\ \midrule
ViT-B    & 14M             & OK-VQA    &  29.45          \\
ViT-B    & 14M+DANCE(part)  & OK-VQA    &  37.56           \\
\midrule
ViT-L    & 14M             & OK-VQA    &  33.14          \\
ViT-L    & 14M+DANCE(part)  & OK-VQA    &  38.55           \\
ViT-L    & 14M+DANCE(whole) & OK-VQA    &  \textbf{39.25}          \\ \bottomrule
\end{tabular}}
\vspace{-0.9em}
\caption{Effect of DANCE for pre-training, testing on existing commonscene benchmark OK-VQA.}
\vspace{-1.1em}
\label{tab:VLCRpretrainOKVQA}
\end{table}

\begin{figure}[t]
  \centering
   \includegraphics[width=0.85\linewidth]{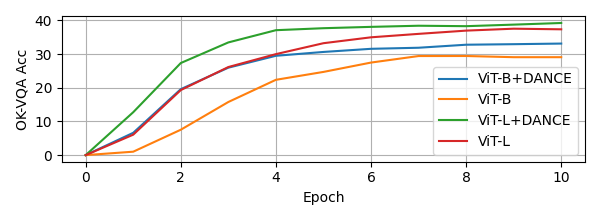}
   \vspace{-1.6em}
   \caption{Performance on existing commonscene benchmark OK-VQA during fine-tuning.}\label{fig:VLCRpretrainOKVQA}
   \vspace{-1em}
\end{figure}

The results are shown in~\Tref{tab:VLCRpretrainOKVQA}. Significant improvements can be seen by comparing the corresponding baselines and DANCE. When DANCE is applied just on part of pre-training data, the performances of ViT-Base and ViT-Large are increased by 3.69\% and 1\% respectively. And by using DANCE on the entire pre-train data, we achieve 1.69\% improvement with ViT-Large.
Besides, as shown in~\Fref{fig:VLCRpretrainOKVQA}, we found that the model pre-trained on DANCE can achieve faster and more stable convergence than the ordinary pre-trained model on the OK-VQA benchmark.

\noindent{\fontsize{10.5}{12}{\textbf{DANCE for Fine-tuning.}}}
DANCE can also contribute positively in the fine-tuning stage. We evaluate the commonsense ability of the model on the proposed diagnostic test set, rather than on existing commonsense knowledge benchmarks, since the latter cannot evaluate models that are not fine-tuned on the specific downstream tasks, e.g., VQA. For fairness, DANCE and the corresponding baseline are fine-tuned on the same number of batches and steps, and on the same training set of COCO images.
The results are reported in the last two rows of ~\Tref{tab:VLCRpretrainfinetune}. Compared with the baseline, DANCE fine-tuning brings significant improvement in the commonsense test sets, and can still obtain comparable results on COCO.

\begin{table}[t]
\setlength{\tabcolsep}{0.7mm}
\centering
\scalebox{0.95}{
\begin{tabular}{@{}c|ccccc@{}}
\toprule
DANCE Ratio  & 50\%    & 30\%    & 10\%    & 0\%     & 50$\rightarrow$10\% \\ \midrule
OK-VQA Acc & 30.32 & 32.48 & 32.03 & 29.45 & \textbf{33.14}                               \\ \bottomrule
\end{tabular}}
\vspace{-1em}
\caption{Ablation of proportions of DANCE-augmented data.}
\vspace{-1.5em}
\label{tab:AblationRatio}
\end{table}

\noindent{\fontsize{10.5}{12}{\textbf{Ablation.}}}
In the table~\ref{tab:AblationRatio}, we study the impact of different proportions of DANCE-augmented data, i.e., $p$, on the performance in the downstream OK-VQA task. We find that a suitable ratio (30\%) is beneficial to the performance improvement, while too high (50\%) or too low (10\%) reduces the performance improvement, and the curriculum learning strategy that linearly decreases the ratio from 50\% to 10\% achieves the best performance. The experiments are based on ViT-B backbone and 14M pre-train data.

\section{Conclusion and Future Work}
This paper takes a step towards injecting commonsense capability into VL-models.
We first observed that VL-models are lacking commonsense ability as existing popular VL datasets do not contain much commonsense knowledge. Therefore, we propose a new training strategy DANCE which is compatible with most VL-models, by training on images paired with our generated entity-hidden commonsense riddles, in a scalable and automatic way. To support the commonsense evaluation of a suite of VL-models in a well-received way, a retrieval-based commonsense diagnostic benchmark is built. We then empirically verify the weaknesses of existing VL-models and the effectiveness of DANCE.
Despite significant improvements in both the existing commonsense and our diagnostic benchmarks, we still face challenges. 
Towards human-like intelligence, awareness of commonsense knowledge is not enough. The model should be able to do reasoning, such as mathematical and physical calculations in real-life scenarios. This is still weak in existing VL-models and is not included in existing commonsense knowledge bases. Future research could be conducted to analyze and improve various reasoning aspects of VL-models.

{\small
\bibliographystyle{ieee_fullname}
\bibliography{egbib}
}

\clearpage

This appendix is organized as follows:
\begin{itemize}
    \item In~\Sref{VLvsNLP}, we further illustrate the commonsense lacking issue by providing additional comparison of fundamental VL datasets with commonly used NLP data.
    \item In~\Sref{quali}, we provide more visualizations of success examples of our method on the proposed diagnostic benchmark for both text-image and image-text retrieval.
    \item In~\Sref{quali-okvqa}, we provide more visualizations of success examples of our method on the OK-VQA benchmark.
    \item In~\Sref{stat-testset}, we summarize the statistics of the proposed diagnostic test data.
    \item In~\Sref{supp-failurecase}, we study the failure case of our DANCE augmented model.
\end{itemize}

\section{Commonsense in Fundamental VL Data vs NLP Data}\label{VLvsNLP}

\begin{figure*}[t]
  \centering
   \includegraphics[width=0.65\linewidth]{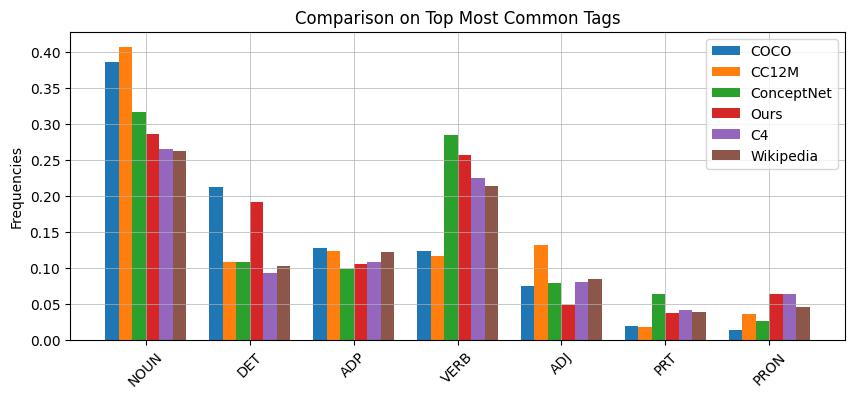}
   \includegraphics[width=0.65\linewidth]{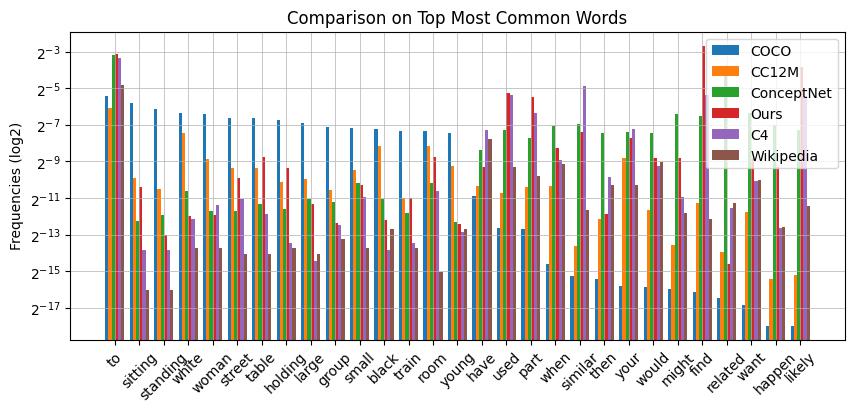}
   \caption{Comparison of the syntactic categories and words distributions of fundamental VL data (COCO~\cite{lin2014microsoft} and CC12M~\cite{changpinyo2021conceptual}), ours training data generated by DANCE, and commonly used NLP data (ConceptNet~\cite{speer2017conceptnet}, Wikipedia~\cite{wikidump} and C4~\cite{2019t5}). Commonsense is lacking in VL data compared with NLP data, and is improved by DANCE strategy.}\label{fig:supp-VLvsNLP}
\end{figure*}

We further explore the commonsense lacking issue in the current fundamental VL data by comparing them with common natural language processing (NLP) data. Here we compare the distributions of the syntactic categories and words of the most popular VL datasets (COCO~\cite{lin2014microsoft} and CC 12M~\cite{changpinyo2021conceptual}) with three commonly used NLP datasets: ConceptNet~\cite{speer2017conceptnet} the knowledge base dataset, Wikipedia~\cite{wikidump} the popular~\cite{devlin2018bert,liu2019roberta,Sanh2019DistilBERTAD,DBLP:journals/corr/abs-1909-11942,DBLP:journals/corr/abs-1906-08237} cleaned English-language articles with the size of $16$GB, C4~\cite{2019t5} the popular used~\cite{DBLP:journals/corr/abs-2105-03824,https://doi.org/10.48550/arxiv.2101.03961,narang2021transformer,tay2021scale,tay2021charformer,kim2021bert,scao2022language} English-language text sourced from the Common Crawl web scrape with the size of $745$GB. The syntactic categories and word distributions comparison is shown in~\Fref{fig:supp-VLvsNLP}.

The upper part of \Fref{fig:supp-VLvsNLP} shows the distribution of the most frequent part-of-speech (POS) tags with punctuation marks excluded, and the lower part shows the most frequent word tokens. There is a significant difference between top POS tag/word token distributions of VL datasets compared with those of the regular texts. Similar to our observation in the main paper, the most frequent words in the text in existing VL datasets are nouns (NOUN) for \textbf{individual entities}, like ``\textit{street}'', ``\textit{table}'', ``\textit{train}''. In contrast, all the NLP datasets have apparently more verbs (VERB), like ``\textit{have}'', ``\textit{used}'', ``\textit{find}'', ``\textit{want}'', ``\textit{happen}'' that contains richer information about the \textbf{relationship between entities}. Besides, the NLP datasets include more particles (PRT), like ``\textit{to}'', and pronouns (PRON) like ``\textit{your}'', which are associated with \textbf{interconnection} information. This further illustrates the lacking commonsense issue in the fundamental VL datasets. 

While the implicit information about the \textbf{interconnections between entities} is in high demand for developing commonsense and reasoning ability, the fundamental VL datasets are lacking it. This motivates us to use commonsense knowledge to improve VL data. In addition, the distribution of ours training data is also included for comparison. We can see that our data is similar to NLP data in terms of the interconnection between entities.

\section{Additional Qualitative Results on Our Diagnostic Benchmark}\label{quali}

\begin{figure*}[t]
  \centering
   \includegraphics[width=0.85\linewidth]{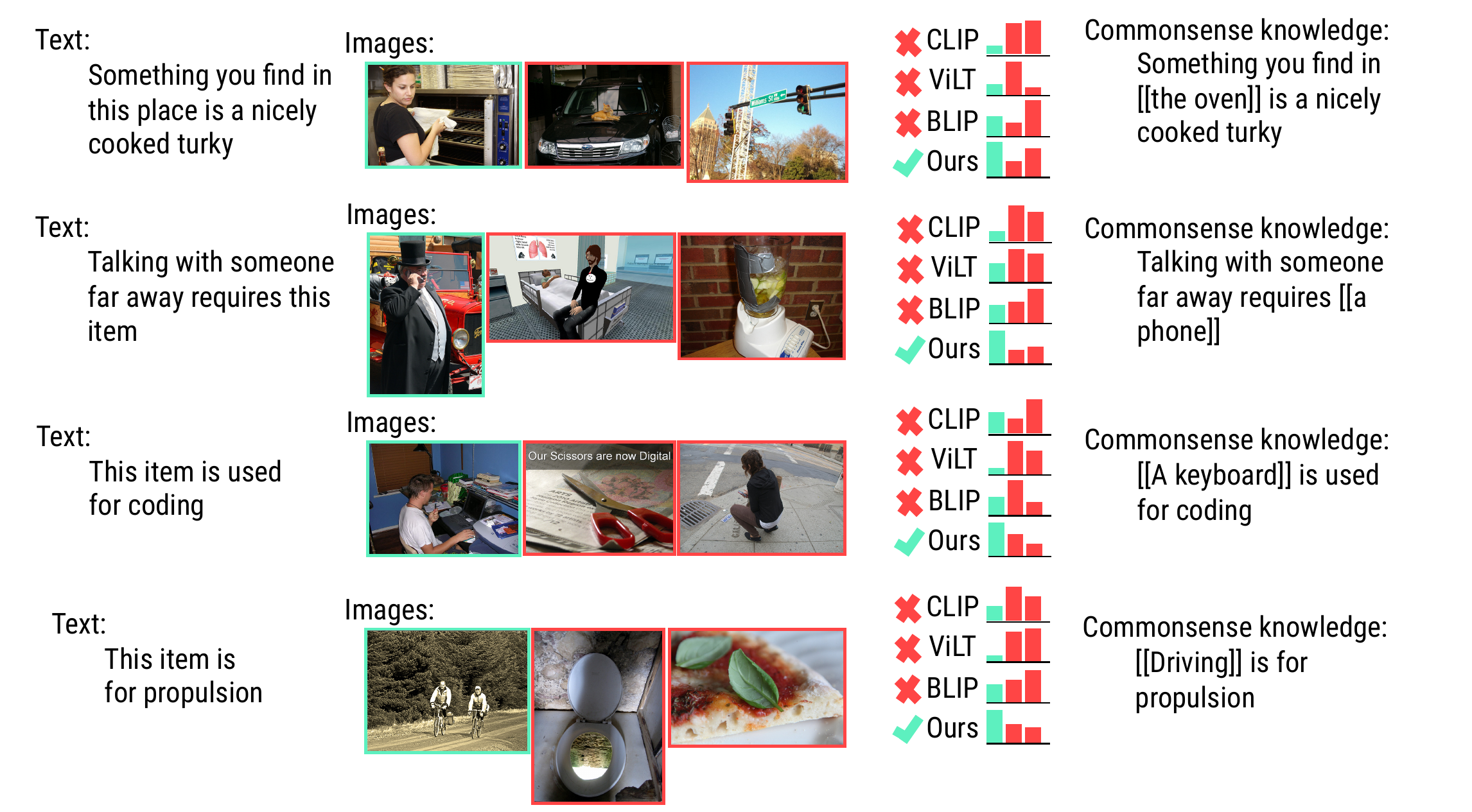}
   \caption{Qualitative examples from our diagnostic test set for text-image retrieval.}\label{fig:supp-quali-TI}
\end{figure*}

\begin{figure*}[t]
  \centering
   \includegraphics[width=0.85\linewidth]{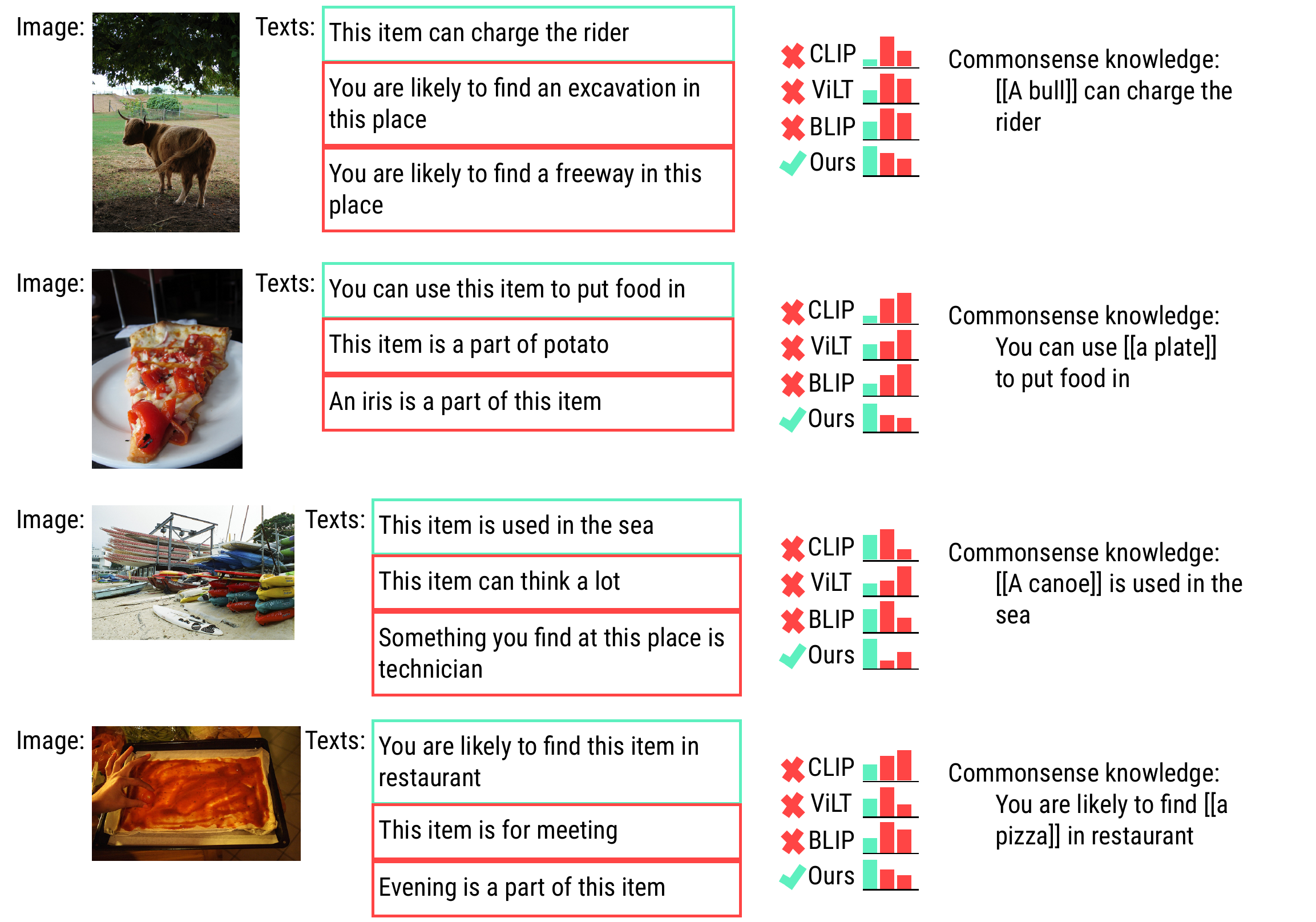}
   \caption{Qualitative examples from our diagnostic test set for image-text retrieval.}\label{fig:supp-quali-IT}
\end{figure*}

In~\Fref{fig:supp-quali-TI} and~\Fref{fig:supp-quali-IT}, we show additional qualitative comparison with the state-of-the-art VL-models on our diagnostic test set for text-image and image-text retrieval respectively. In~\Fref{fig:supp-quali-TI}, from left to right is the input text, the input images including a correct one (in blue) and two incorrect ones (in red), the scores by each individual model, and the commonsense knowledge from the knowledge graph~\cite{speer2017conceptnet} that required for retrieval.
In~\Fref{fig:supp-quali-IT}, from left to right is the input image, the input texts including a correct one (in blue) and two incorrect ones (in red), the scores, and the related commonsense knowledge from the knowledge graph. We can see that all the baselines fail to identify the correct answers, which further illustrates the lacking of commonsense ability in the popular VL-models. In contrast, our DANCE pre-trained model successfully retrieves the correct ones. We note that all these images and the knowledge are held out from the training set. This further demonstrates the reasoning ability enhanced by our DANCE strategy.

\section{Additional Qualitative Results on OK-VQA Benchmark}\label{quali-okvqa}

\begin{figure*}[t]
  \centering
   \includegraphics[width=0.89\linewidth]{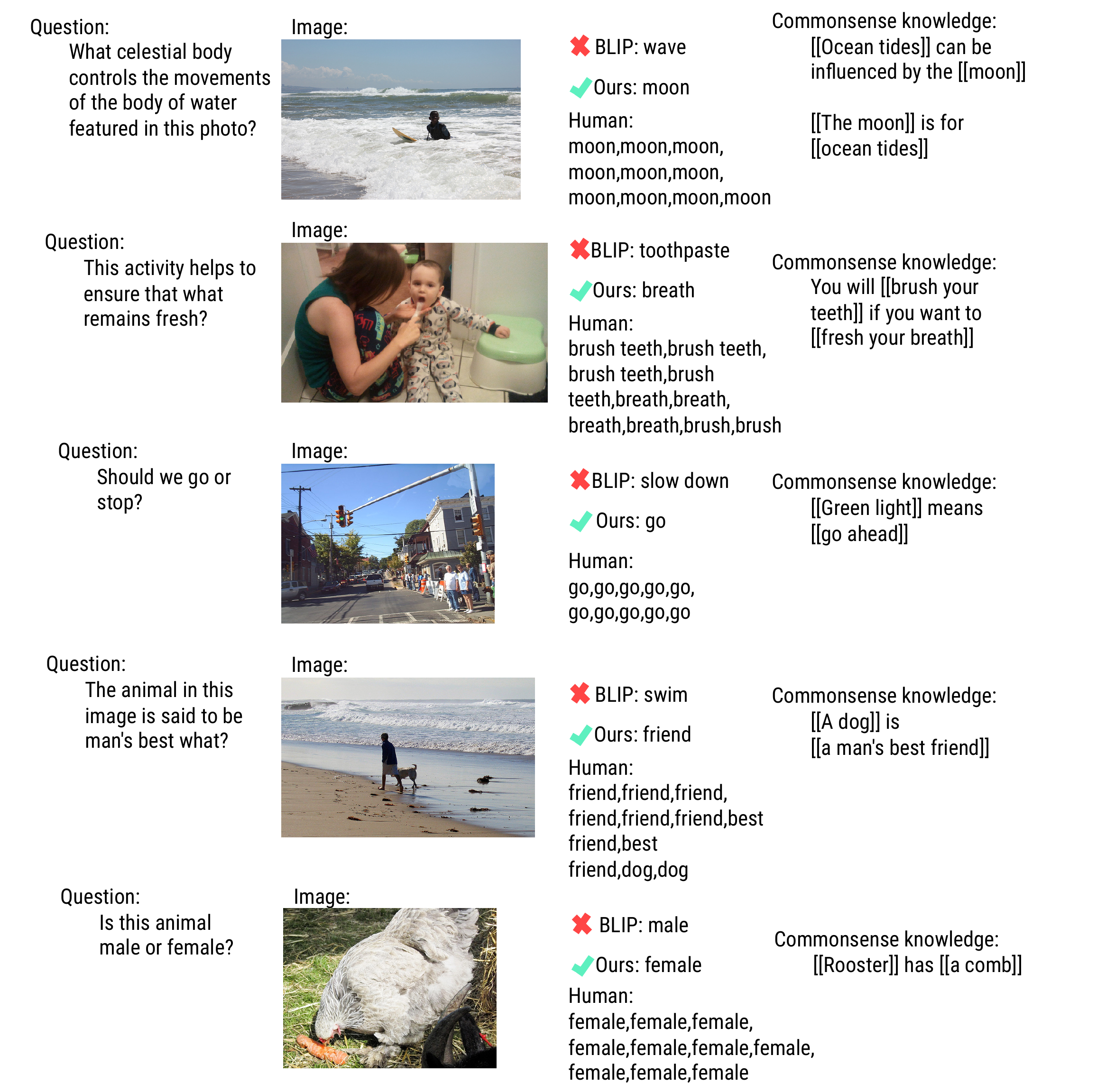}
   \caption{Qualitative examples from the commonsense-aware benchmark OK-VQA.}\label{fig:supp-quali-okvqa}
\end{figure*}

In~\Fref{fig:supp-quali-okvqa}, we show additional qualitative comparison with the state-of-the-art VL-models on the official validation split of the popular commonsense-aware OK-VQA dataset. We note that the validation split is not included during fine-tuning. From left to right is the input question, the input image, the answers by the baseline model BLIP, the DANCE pre-trained model and human, and the related commonsense knowledge from the knowledge graph.
The baseline model struggles with these questions and predicts some relevant but wrong answers, which further demonstrates the lack of commonsense ability in the current VL-models. DANCE improves the VL-model's commonsense ability in numerous aspects, including the commonsense knowledge of physics as shown in the first row, the commonsense of human behavior and motivation in the second and third rows, and the knowledge about animals in the fourth and fifth rows.
This further demonstrates the commonsense ability enhanced by our DANCE strategy.

\section{Statistics of Our Diagnostic Benchmark}\label{stat-testset}

\begin{table*}[t]
\centering
\begin{tabular}{@{}c|cccc@{}}
\toprule
          & Text-Image seen & Text-Image unseen & Image-Text seen & Image-Text unseen \\ \midrule
\# Images & 4949            & 4974              & 500             & 500               \\
\# Texts  & 500             & 500               & 13930           & 14889            \\
\# Seen Images & 0 & 0 & 0 & 0 \\
\# Seen Texts & 500 & 0 & 13930 & 0 \\
\bottomrule
\end{tabular}
\caption{Statistics of different splits of our diagnostic benchmark.}\label{table:supp-stat}
\end{table*}

In~\Tref{table:supp-stat}, we show the statistics of the four different splits of our diagnostic retrieval test set. Each row respectively represents the number of different images, the number of different text or riddles in each split, and the number of different images and texts that also appear in the training data. All these images for our test set does not appear in the training set. The knowledge in both Text-Image unseen split and Image-Text unseen split is held out from the training set.

\section{Failure Case on OK-VQA Benchmark}\label{supp-failurecase}

\begin{figure*}[t]
  \centering
   \includegraphics[width=0.65\linewidth]{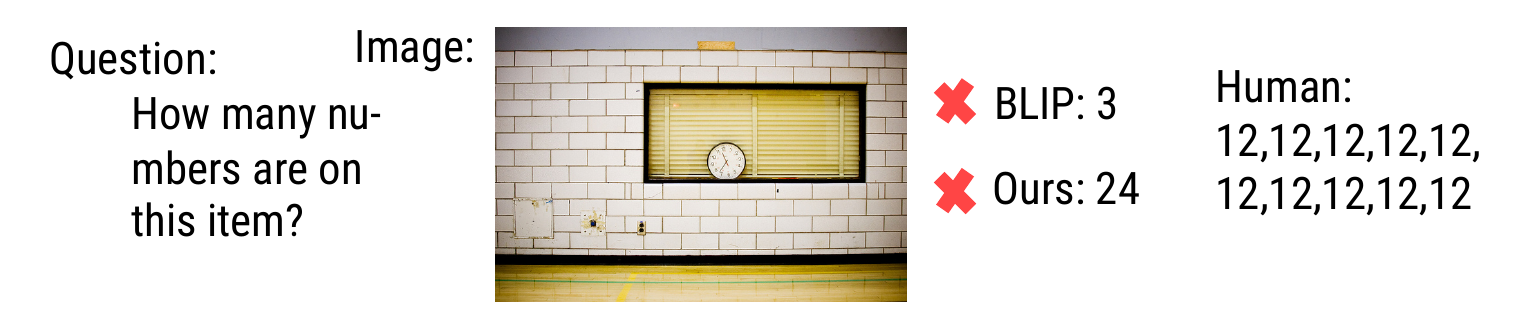}
   \caption{Case study of failure on the OK-VQA benchmark.}\label{fig:supp-failcase}
\end{figure*}

In the main paper, we mainly focus on enhancing the VL-model's ability to general commonsense via combining the VL data lacking commonsense with commonsense knowledge graphs. However, our model learned from this commonsense-augmented data still suffers in some special real-life scenarios. Here we visualize the failure case of the model with DANCE pre-training in~\Fref{fig:supp-failcase}. The model fails to answer a question about counting or quantity. This indicates that the sense of numbers or the mathematical reasoning ability is still weak in existing VL-models, which is also not included in existing commonsense knowledge bases.

\end{document}